\pdfoutput=1

\documentclass[11pt]{article}

\usepackage{acl}
\usepackage{pifont}
\usepackage{times}
\usepackage{latexsym}
\usepackage{fontawesome}
\usepackage[T1]{fontenc}
\usepackage{marvosym}
\usepackage{hyperref}
\usepackage{enumitem}
\usepackage{amsmath}
\usepackage{algorithmic}
\usepackage{algorithm}
\usepackage{tcolorbox}
\usepackage{colortbl}
\usepackage{booktabs}
\usepackage{multirow}
\usepackage{makecell}
\usepackage{balance}
\usepackage{soul, color, xcolor}
\usepackage{caption}

\usepackage[utf8]{inputenc}

\usepackage{microtype}

\usepackage{inconsolata}
\usepackage{graphicx}
\usepackage{marvosym}
\usepackage{amssymb}

\title{\textsc{CLI-RAG}: A Retrieval-Augmented Framework for Clinically Structured and Context Aware Text Generation with LLMs}

\author{Garapati Keerthana$^{1}$, Manik Gupta$^{1}$ \\
$^{1}$Birla Institute of Technology and Science, Pilani, Hyderabad, India \\  
 \texttt{\{p20240505,manik\}@hyderabad.bits-pilani.ac.in}
}

\begin{document}
\maketitle
\begin{abstract}
Large language models (LLMs), including zero-shot and few-shot paradigms, have shown promising capabilities in clinical text generation. However, real-world applications face two key challenges: (1) patient data is highly unstructured, heterogeneous, and scattered across multiple note types; and (2) clinical notes are often long and semantically dense, making naive prompting infeasible due to context length constraints and the risk of omitting clinically relevant information.

We introduce \textsc{CLI-RAG} (\underline{C}linically \underline{I}nformed \underline{R}etrieval-\underline{A}ugmented \underline{G}eneration), a domain-specific framework for structured and clinically grounded text generation using LLMs. It incorporates a novel hierarchical chunking strategy that respects clinical document structure and introduces a task-specific dual-stage retrieval mechanism. The global stage identifies relevant note types using evidence-based queries, while the local stage extracts high-value content within those notes creating relevance at both document and section levels.

We apply the system to generate structured progress notes for individual hospital visits using 15 clinical note types from the MIMIC-III dataset. Experiments show that it preserves temporal and semantic alignment across visits, achieving an average alignment score of 87.7\%, surpassing the 80.7\% baseline from real clinician-authored notes. The generated outputs also demonstrate high consistency across LLMs, reinforcing deterministic behavior essential for reproducibility, reliability, and clinical trust.
\end{abstract}

\section{Introduction}

Large language models (LLMs) have shown remarkable success in natural language processing \cite{DBLP:openai2023gpt4, yang2024qwen2}, including clinical applications such as summarization \cite{agrawal-etal-2022-large, wang2023chatcad}, medical Q\&A \cite{singhal2023large}, and decision support \cite{fang2023medicalexam}. However, deploying LLMs in real-world clinical settings remains challenging due to the unstructured, fragmented, and semantically dense nature of electronic health records (EHRs) \cite{rule2021length, kuhn2015clinical, meystre2008extracting, wang2017characterizing}. Clinical documentation spans multiple heterogeneous note types, e.g. nursing, radiology, consultations, each with varied structure and granularity, often containing redundant or incomplete content \cite{annurev2021, markel2010copy, apathy2022early}.

Progress notes, typically structured in the SOAP format (Subjective, Objective, Assessment, Plan), are essential for ongoing care but are frequently missing in real-world datasets appearing in only 8.56\% of visits in MIMIC-III \cite{johnson2016mimic}. Reconstructing such notes from fragmented sources requires reasoning over multiple documents, temporal alignment, and adherence to clinical structure. Off-the-shelf retrieval-augmented generation (RAG) approaches \cite{lewis2020retrieval, izacard2022few} are ill-suited for this task, as they operate on flat corpora and ignore clinical semantics, note provenance, or task-specific relevance.

We introduce \textsc{CLI-RAG} (\underline{Cl}inically \underline{I}nformed \underline{R}etrieval-\underline{A}ugmented \underline{G}eneration), a structured generation framework designed to synthesize SOAP-format progress notes by composing evidence from diverse EHR notes. Our method addresses two key research questions: 1. What information is most relevant to progress note generation, regardless of its source? 2. Which note types consistently contribute towards this information?

To answer these, our system proposes a dual-stage retrieval strategy. A global retrieval step leverages task-specific clinical queries to extract relevant content across all note types, followed by a local retrieval phase that drills down into note-type-specific content using tailored sub-queries. The pipeline includes clinically structured preprocessing, hierarchical chunking, metadata-guided embeddings, and temporally conditioned prompting to simulate real-world documentation workflows.

We evaluate our system on 1,108 patient visits from MIMIC-III dataset, measuring lexical, semantic, structural, and temporal fidelity. Our system achieves a temporal alignment score of 87.7\%, outperforming clinician-authored notes 80.7\%. These findings highlight the effectiveness of clinically informed retrieval in generating coherent, faithful progress notes, with implications for summarization, documentation support, and synthetic EHR generation.

\section{Methodology}
We present a structured retrieval-augmented generation framework tailored for synthesizing clinical progress notes from heterogeneous electronic health record (EHR) sources. The system is evaluated on the task of generating detailed SOAP-format notes by composing evidence from 15 diverse note types in the MIMIC-III dataset \cite{johnson2016mimic}. Unlike generic RAG pipelines that treat documents uniformly, our approach models the hierarchical structure and clinical semantics inherent in multi-source clinical documentation to produce temporally coherent and clinically grounded outputs.
\subsection{Overview of \textsc{CLI-RAG}}
\begin{figure*}[ht]
    \centering
    \includegraphics[width=1.0\textwidth]{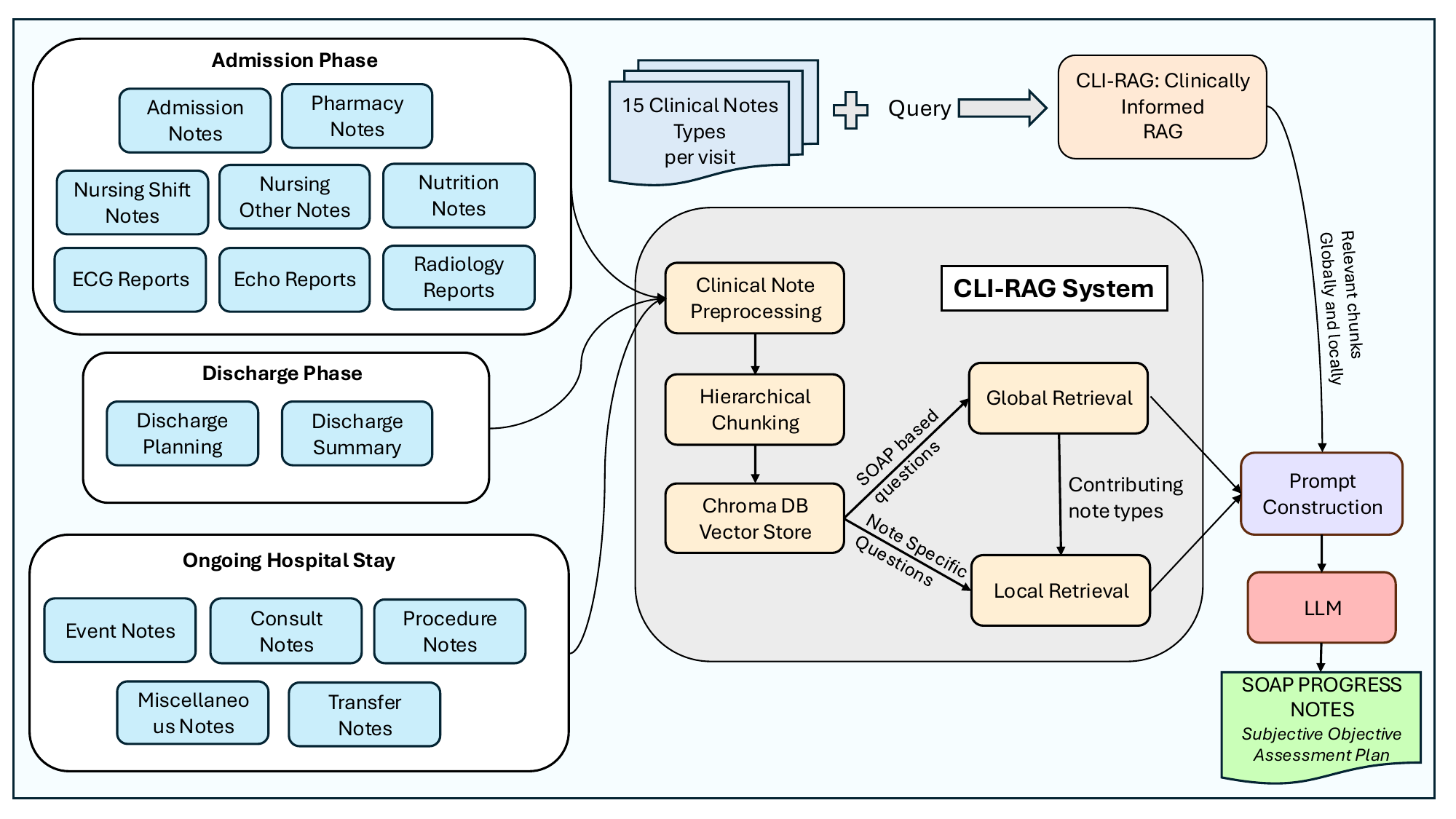}
    \caption{Overview of the \textsc{CLI-RAG} framework: Fifteen structured clinical note types from each hospital visit serve as evidence sources. These are preprocessed, hierarchically chunked, and passed through a dual-stage retrieval pipeline before constructing prompts for large language models to generate structured progress notes.}
    \label{fig:cli_rag_overview}
\end{figure*}

Our framework is centered around a dual-stage, clinically informed retrieval architecture that addresses two core challenges in clinical NLP: (1) fragmented information across heterogeneous note types, and (2) long, unstructured text that often exceeds the context window of large language models (LLMs). To address this, we employ a global retrieval step to identify high-value note types using predefined clinical questions, followed by local retrieval to extract fine-grained evidence from specific sections within those notes.
Notes are preprocessed and segmented into hierarchical chunks that preserve document structure, encoded using the all-mpnet-base-v2 sentence transformer \cite{reimers-2019-sentence-bert}, and indexed with ChromaDB\footnote{\url{https://www.trychroma.com/}} for efficient semantic retrieval. At inference, retrieved chunks are de-duplicated, reranked, and assembled into prompts with structured metadata. For longitudinal consistency, summaries of prior visits are optionally incorporated.
This modular pipeline enables the generation of clinically faithful, temporally coherent progress notes. An end-to-end architecture and flow is shown in Figure~\ref{fig:cli_rag_overview}.

\subsection{Preprocessing and Hierarchical Chunking}
Raw clinical notes in electronic health records (EHRs) are rife with inconsistencies: they include de-identification artifacts, UI-generated noise, variable casing, non-standard bullets, and redundant line breaks. Before any information retrieval or language generation step, we perform a robust preprocessing pipeline aimed at denoising, normalizing, and structurally realigning clinical text into a form conducive to semantic understanding.
The preprocessing module executes a cascade of transformations: it replaces Unicode symbols, removes [**DEID**] tokens while preserving their content where safe, strips JavaScript artifacts from the user interface, normalizes whitespace and line breaks, and collapses common bulleting styles into canonical formats. Additionally, numerical blocks such as vitals or lab values are heuristically converted into key-value format (e.g., \texttt{WBC} $\rightarrow$ \texttt{WBC: 17.5}) to preserve measurement semantics across lines. Finally, all-caps section headers are normalized to title case and missing colons are inserted when necessary. This stage ensures that textual content, often spanning multiple note types and documentation systems, becomes structurally comparable.

Beyond text normalization, we introduce a novel \textbf{hierarchical chunking strategy} grounded in clinical structure. Instead of naively chunking based on fixed-length windows, our system respects the logical organization of clinical documents. Each note is first segmented at the granularity of high-level clinical section headers such as \texttt{History of Present Illness}, \texttt{Assessment}, or \texttt{Impression} which reflect discrete semantic zones of documentation. These header-segmented blocks are then recursively divided into smaller sub-chunks based on character length thresholds with fixed overlaps, ensuring downstream retrievability within embedding limits, while maintaining contextual continuity.
Unlike flat chunking, this approach preserves intra-document cohesion, enables section-aware retrieval, and dramatically reduces boundary-related semantic leakage during encoding. For instance, two separate chunks from an \texttt{Assessment and Plan} section will be co-located in retrieval space, while still being individually embeddable. Each chunk is then indexed along with metadata identifying its patient ID, hospital admission ID, visit date, originating note type, clinical section, and local chunk identifier that tracks the count within the clinical section as this can be further used to order the chunks to maintain original clinical context ordering during retrieval and this forms the foundational unit of knowledge for downstream retrieval.
Given the variability in clinician documentation styles, not all segments of a note will cleanly match a known section. To accommodate this, we implemented a fallback mechanism where unmatched spans are tagged under an \texttt{Unlabeled} section during the chunking stage. These \texttt{Unlabeled} chunks preserve full contextual information and are treated as first-class entities, assigned their own \texttt{chunk\_id} for ordering. During retrieval and prompt construction, any \texttt{Unlabeled} content deemed relevant is explicitly included and annotated as \textit{``More Information''} within the final prompt structure, signaling to the language model that these passages may contain clinically useful but unstructured context. This design choice ensures that no potentially valuable information is discarded purely due to formatting inconsistency, and that the model retains access to all relevant cues needed for generation, even when embedded in free-text form.
This structurally aligned pre-processing pipeline shown in Figure~\ref{fig:horizontal-cleaning-pipeline} enables high-fidelity semantic search by mitigating the brittleness of syntactic variance across clinical notes. More importantly, it primes the system for retrieval and generation strategies that are not merely lexical but clinically aware.

\subsection{Multi-Stage Retrieval: Global and Local Contextualization}

To accurately reconstruct clinically grounded progress notes from fragmented EHR narratives, our sytem employ a dual-stage retrieval architecture that integrates both global and local contextual signals. This design reflects a central hypothesis: the information necessary to generate comprehensive progress notes is scattered across multiple note types and sections, and must first be surfaced and contextualized in a structured manner.

\paragraph{Global Retrieval.} The first stage in the retrieval pipeline operates at the \textit{visit level} and is tasked with identifying globally relevant clinical information regardless of where it appears in the patient record. Given a predefined set of task-driven global clinical questions (e.g., \textit{``What symptoms did the patient report?''}, \textit{``What procedures were performed?''}), the system performs dense retrieval over all available notes for a patient visit using sentence-level embeddings. For each question, a semantic query embedding is generated and a similarity search is performed over a vector index built from all chunked clinical documentation corresponding to the same \texttt{patient\_id} and \texttt{visit\_date}. Crucially, no constraints are placed on the \texttt{note\_type} dimension at this stage allowing the system to retrieve across all 15 clinical document categories. This wide filtering enables maximal coverage and provides an unbiased scan of what information is available and from where.

Retrieved chunks are then passed through a deduplication module, which filters out highly similar spans using pairwise cosine similarity thresholds to avoid redundant inclusion of semantically duplicated information across note types (e.g., similar text copied between a nursing and physician note). To improve retrieval fidelity, we use a hybrid scoring approach that linearly combines BM25 lexical relevance and embedding-based cosine similarity. This ensures that retrieved evidence is both semantically aligned and lexically grounded. Finally, symptom-focused reranking prioritizes chunks likely to contain clinically salient observations using hand-crafted symptom keyword heuristics.

This global retrieval phase answers two key research questions: (1) What pieces of clinical information are most relevant for downstream progress note generation, regardless of source note type? and (2) Which note types contribute most frequently to these task-relevant evidence spans? The system logs metadata about retrieved evidence chunks, including their note type and section, enabling precise tracking of which document types consistently support specific clinical tasks. This allows us to both generate structured notes and introspectively audit which sources contribute meaningfully to clinical synthesis.

\paragraph{Local Retrieval.} Having identified the most informative note types through global retrieval, our system proceeds to a second stage: fine-grained retrieval within individual note types. For each contributing note type, a curated set of local questions is queried (e.g., for \texttt{progress\_notes}, we ask \textit{``What is today’s clinical assessment?''}, \textit{``Were ABG results reported?''}). This ensures that task-relevant content is retrieved not only across notes but within the structural logic of individual note types.

Unlike the global phase, local retrieval restricts the search space not just by \texttt{patient\_id} and \texttt{visit\_date}, but also by \texttt{note\_type}. This additional constraint allows the system to isolate semantically dense evidence from specific clinical roles such as radiologist interpretations, ICU nurse logs, or attending physician assessments thus enabling structured extraction tuned to the documentation practices of each source.

Local retrieval is performed using the same hybrid scoring, deduplication, and reranking strategy as the global phase, but scoped only to documents within the specified note type. This constrains the search space and enables highly targeted extraction of contextual content. Importantly, by posing note-type-specific questions, we preserve the original structural intent behind each document (e.g., distinguishing subjective narrative in nursing notes from diagnostic impressions in radiology reports), which enhances the faithfulness of generation.

\paragraph{Retrieval Oriented Metadata.} All retrieved chunks whether from global or local retrieval are embedded with rich metadata, including \texttt{note\_type}, \texttt{header\_name}, \texttt{chunk\_id\_in\_header}, and temporal tags. These are used both for reranking and for organizing content during prompt construction. In cases where a chunk does not match any known clinical section from our exhaustive header set, it is labeled as \texttt{Unlabeled} and later surfaced as \textit{``More Information''} during prompt formatting. This design ensures that valuable but structurally ambiguous content is not discarded, but rather explicitly marked for LLM consumption.

\paragraph{Clinical Semantics via Retrieval.} This dual-stage retrieval strategy provides a principled mechanism to structure evidence-driven prompting. The global stage captures macro-level salience by identifying relevant note types and tasks, while the local stage enables fine-grained contextualization. Together, they simulate a clinical reasoning workflow: identifying where to look, and then extracting what to say. Retrieval thus becomes not only a retrieval operation, but a semantic alignment step that translates raw EHR sprawl into task-aligned, evidence-rich scaffolds for generation. The entire pipeline is depicted in Appendix~\ref{fig:cli_rag_pipeline}

\paragraph{Prompt Construction and LLM-Based Generation.} Once relevant context has been retrieved through global and local passes, we construct an inference prompt for downstream generation of structured clinical progress notes. Retrieved chunks are sorted by \texttt{note\_type}, \texttt{section header}, and \texttt{chunk ID}, preserving original documentation structure and temporal order. Each chunk includes standardized metadata brackets (e.g., \texttt{[progress\_notes | Assessment | chunk 2]}) to anchor the source and enable interpretability. Chunks labeled as \texttt{Unlabeled} during preprocessing are explicitly surfaced under a separate section titled \texttt{More Information}, ensuring they remain accessible during generation despite lacking section headers.

A task-specific prompt template guides the LLM to generate notes in the canonical SOAP format (\textbf{Subjective}, \textbf{Objective}, \textbf{Assessment}, \textbf{Plan}), while enforcing factual grounding. The instruction emphasizes clinical fidelity, discourages hallucination, and encourages structured reasoning using only the extracted content. When generating longitudinal notes (for visits beyond the first), the prior note is summarized and included to help the model distinguish new findings and reason about temporal evolution. This modular prompt structure, coupled with chunk-level ordering and scoped summarization, allows the model to produce detailed, coherent, and reproducible progress notes aligned with clinical expectations.

\section{Experimental Setup and Evaluation}

We evaluate our system across multiple axes lexical fidelity, semantic alignment, structural adherence, and temporal consistency using two open-source LLMs: \textbf{LLaMA-3 70B} and \textbf{Mistral-7B}. Both models generate SOAP-style progress notes grounded in multi-note evidence per hospital visit. We benchmark these outputs against clinician authored progress notes from the MIMIC-III dataset \cite{johnson2016mimic}.

\subsection{Cohort and Setup}

We curated 56 patients with 10–57 visits each, totaling 1,108 hospital encounters. For each visit, a progress note was generated using contemporaneous documentation (excluding progress notes). For longitudinal realism, prior visit summaries were included for all visits beyond the first.

\subsection{Evaluation Dimensions}

\paragraph{Lexical Similarity.} We compute BLEU and ROUGE (1/2/L) to capture token-level and n-gram overlap. As expected in paraphrastic clinical text, scores are low across both models.
\paragraph{Semantic Alignment.} Cosine similarity between embeddings (from \texttt{all-mpnet-base-v2}) measures meaning preservation across generated and gold notes.
\paragraph{Structural Completeness.} Each note is checked for the canonical SOAP sections. Both models consistently yield full coverage of Subjective, Objective, Assessment, and Plan.
\paragraph{Length Control.} Length ratio (generated to gold) evaluates verbosity alignment. Both models produce slightly longer but controlled outputs.
\paragraph{Temporal Coherence.} For longitudinal alignment, cosine similarity between adjacent notes within a patient trajectory is averaged. Generated notes consistently show stronger temporal consistency than clinician-authored baselines.
\paragraph{LLM Preference Voting.} In a blinded quality check using Mistral-7B, generated notes were preferred over clinician-authored notes in 96\% of cases across dimensions of structure, fluency, and completeness.

\begin{table}[h]
\centering
\scriptsize
\begin{tabular}{l|cc}
\toprule
\textbf{Metric} & \textbf{LLaMA-3 70B} & \textbf{Mistral-7B} \\
\midrule
BLEU & 0.0116 & 0.0004 \\
ROUGE-1 & 0.2738 & 0.1783 \\
ROUGE-2 & 0.0743 & 0.0312 \\
ROUGE-L & 0.1102 & 0.0676 \\
Semantic Similarity & 0.7398 & 0.7511 \\
SOAP Sections Present & 4.00 & 4.00 \\
Length Ratio & 1.1975 & N/A \\
Temporal Consistency (Gold) & 0.807 & 0.9126 \\
Temporal Consistency (Generated) & 0.877 & 0.8248 \\
Alignment Ratio & 1.089 & 0.9038 \\
LLM Voting Preference & 96\% & 93.5\% \\
\bottomrule
\end{tabular}
\caption{Evaluation metrics across LLaMA-3 70B and Mistral-7B on 56-patient cohort. Semantic similarity and structural adherence are consistently strong; temporal coherence exceeds real notes.}
\label{tab:metric_summary}
\end{table}

\subsection{Interpretation}

Despite low lexical overlap a known artifact in clinical generation semantic similarity remains high across both models, with SOAP section coverage perfectly retained. LLaMA-3 showed higher temporal alignment (0.877) and richer lexical diversity (higher ROUGE), whereas Mistral-7B produced more concise notes with stronger gold alignment (0.9126 vs. 0.807). Importantly, both models outperformed real notes in longitudinal coherence.

These results highlight the robustness and transferability of our retrieval-augmented approach across LLMs. Regardless of model size or decoding variance, the system produces notes that are clinically structured, semantically faithful, and temporally aligned.

\section{Related Work}
\paragraph{Clinical Note Generation.}
Large language models (LLMs) have demonstrated strong capabilities in clinical text generation, including discharge summaries \cite{wang2023chatcad, williams2024, ellershaw2024automated}, SOAP-format notes \cite{soni2024toward}, and dialogue-informed documentation \cite{biswas2024intelligent}. However, most methods operate over single-note or single-visit inputs without integrating heterogeneous sources or modeling longitudinal clinical reasoning. Encoder–decoder approaches like ClinicalT5 \cite{lu2022clinicalt5} and MedSum \cite{tang-etal-2023-gersteinlab} target isolated summarization tasks but lack mechanisms for cross-visit synthesis or structure-aware generation.
\paragraph{Retrieval-Augmented Generation in Healthcare.}
Retrieval-Augmented Generation (RAG) techniques have shown utility in open-domain QA and summarization by injecting external evidence \cite{lewis2020retrieval, izacard2022few, guu2020retrieval}. Clinical RAG applications have largely focused on static resource grounding \cite{sohn-etal-2025-rationale, Van_Veen_2024} or document-level retrieval from knowledge bases. Few systems explore context-aware retrieval from longitudinal EHRs, and even fewer account for clinical sectioning, document metadata, or temporal relevance. Our system introduces a domain-adapted RAG architecture that bridges these gaps.
\paragraph{Longitudinal and Multi-source Modeling.}
Modeling patient trajectories over time is critical in clinical NLP but remains underexplored in generative settings. Prior efforts in structured disease modeling \cite{dieng2019dynamic, isonuma-etal-2020-tree}, patient timeline summarization \cite{jain2022surveymedicaldocumentsummarization}, and temporal QA \cite{shimizu-etal-2024-qa} provide valuable insights, but most do not synthesize visit-specific notes across multiple modalities. Our approach conditions generation on temporally evolving patient context, explicitly incorporating past visit summaries and diverse note types to improve coherence.
\paragraph{Evaluation of Clinical Generation.} Standard lexical metrics like BLEU and ROUGE remain common \cite{lin-2004-rouge}, though they often fail to capture semantic fidelity in paraphrased clinical outputs. Embedding-based similarity \cite{reimers-2019-sentence-bert}, structure-level checks (e.g., SOAP adherence), and temporal alignment across visits offer more robust evaluation. Few prior works jointly assess these dimensions; our evaluation protocol is designed to reflect semantic, structural, and longitudinal fidelity essential for real-world deployment.

\section{Discussion}
This work demonstrates how structured retrieval and longitudinal conditioning can be leveraged to address fragmentation in electronic health records. The proposed dual-stage retrieval pipeline combining global question-driven relevance with note-type-specific local refinement enables the system to extract clinically meaningful context and generate complete, structured progress notes. Incorporating summaries of prior visits facilitates temporal consistency, reflected in higher alignment scores than real clinician-authored notes. This suggests the model not only synthesizes per-visit information faithfully but also maintains narrative coherence over time. Beyond generation, this framework has practical implications for both retrospective data curation and real-time clinical support backfilling documentation gaps, augmenting training data for clinical NLP tasks, or assisting clinicians in note drafting. Future extensions should explore factuality calibration, incorporation of multimodal evidence, and broader generalization to underrepresented specialties. Embedding human-in-the-loop evaluation will be essential to ensure safety, trustworthiness, and integration into real-world workflows.

\section{Conclusion}
We introduce a clinically informed, retrieval-augmented generation framework for synthesizing structured progress notes from heterogeneous EHR sources. By combining hierarchical chunking, dual-stage retrieval, and longitudinal prompt conditioning, the system produces outputs that are semantically aligned, structurally complete, and temporally consistent.

Experiments on MIMIC-III demonstrate strong alignment with clinician-authored documentation and improved coherence across hospital visits. These results point to the framework's potential for enhancing clinical documentation, supporting downstream reasoning tasks, and enabling the construction of synthetic, high-fidelity longitudinal EHR narratives.

\bibliography{references}
\bibliographystyle{acl_natbib}

\clearpage
\appendix

\section{Ethics Statement}

Our work introduces CLI-RAG, a clinically informed retrieval-augmented generation system aimed at generating structured and contextually grounded patient progress notes. While our contributions are technical, we acknowledge the high-stakes nature of clinical applications. To mitigate ethical risks:

\begin{itemize}
    \item \textbf{Data Privacy:} All data used is de-identified and complies with HIPAA and related privacy standards. Our experiments are conducted on publicly available datasets (e.g., MIMIC-III), which are approved for research use.
    \item \textbf{Clinical Safety:} CLI-RAG is designed as a research prototype and is not intended for deployment in real-time clinical decision-making without rigorous validation and oversight. It must not be used as a substitute for expert clinical judgment.
    \item \textbf{Bias and Fairness:} We recognize that pretrained language models may inherit biases present in clinical notes, documentation styles, or healthcare systems. We take care to surface relevant content transparently and plan future work to evaluate representation fairness across subpopulations.
    \item \textbf{Transparency:} CLI-RAG prioritizes interpretability through structured chunk retrieval and evidence attribution. All outputs are traceable to their source chunks to support clinician review.
\end{itemize}

We commit to continuing responsible AI practices, including collaborating with clinicians, incorporating human feedback, and aligning with healthcare regulatory norms before any deployment.

\section{Broader Impact Statement}

CLI-RAG demonstrates the promise of retrieval-augmented generation systems in clinical natural language processing (NLP), particularly for assisting clinicians in generating structured, evidence-grounded summaries from complex medical records. We introduce a modular, interpretable, and clinically grounded RAG system that has the potential to improve clinical documentation, healthcare research, and health informatics. The system is designed to reduce the cognitive and administrative burden placed on healthcare providers by synthesizing raw clinical notes into coherent, structured progress notes. In environments where physicians spend significant time on documentation often at the cost of patient interaction such systems can serve as productivity aids, improving both clinician efficiency and documentation quality.

\textbf{Industrial Applications:} In industrial and commercial healthcare settings, CLI-RAG could be integrated into electronic health record (EHR) platforms to assist with real-time note summarization, longitudinal chart review, and audit-ready report generation. Additionally, its architecture lends itself to medical billing support and clinical coding, where structured output aligned with retrieved evidence can enhance compliance and transparency. This is particularly valuable in hospitals and telemedicine platforms facing documentation backlogs or resource constraints.
\begin{itemize}
    \item \textbf{Clinical Documentation Support:} CLI-RAG can enhance electronic health record (EHR) systems by suggesting high-quality progress notes grounded in retrieved content, reducing clinician burnout and improving documentation quality.
    \item \textbf{Clinical Audit and Coding:} Automated summarization aligned with structured evidence may support downstream tasks like medical coding, insurance audits, or billing documentation.
    \item \textbf{Clinical Research Workflows:} Researchers can use CLI-RAG to rapidly extract relevant information across longitudinal records, enabling cohort construction, phenotype extraction, and retrospective studies.
\end{itemize}

\textbf{Research Contributions:} For clinical researchers, CLI-RAG presents a tool for retrospective data extraction and structured cohort analysis. By automating the summarization of complex patient trajectories, researchers can focus on higher-level tasks such as identifying phenotypes, evaluating treatment responses, and constructing case studies. The system's use of fine-grained chunk retrieval allows researchers to validate summaries against traceable source evidence, supporting reproducibility and methodological rigor.
\begin{itemize}
    \item \textbf{Structured Clinical RAG Paradigms:} CLI-RAG provides a blueprint for structured, interpretable RAG in high-stakes domains, bridging retrieval-based QA and summarization in a modular pipeline.
    \item \textbf{Evaluation Frameworks:} By enabling chunk-level traceability, CLI-RAG lays the foundation for more nuanced intrinsic and extrinsic evaluations of clinical language models.
\end{itemize}
From an NLP perspective, CLI-RAG advances the field by demonstrating how retrieval-augmented generation can be tightly coupled with domain-specific structure and explainability. It bridges traditionally separate tasks such as information retrieval, clinical summarization, and question answering. The framework also sets the stage for new evaluation paradigms, emphasizing interpretability and attribution over black-box generation.

\textbf{Limitations and Responsible Use:} 
While our system shows promise, it must be rigorously validated across diverse clinical environments. Further research is needed to assess generalizability, temporal robustness, and human-AI collaboration in clinical settings. We envision CLI-RAG as a step toward safer, more explainable AI systems for healthcare. We view CLI-RAG not as a final product, but as a foundational system that invites interdisciplinary development, robust clinical testing, and ethical guardrails for future healthcare NLP applications.

\begin{figure*}[t]
    \centering
    \includegraphics[width=1.0\textwidth]{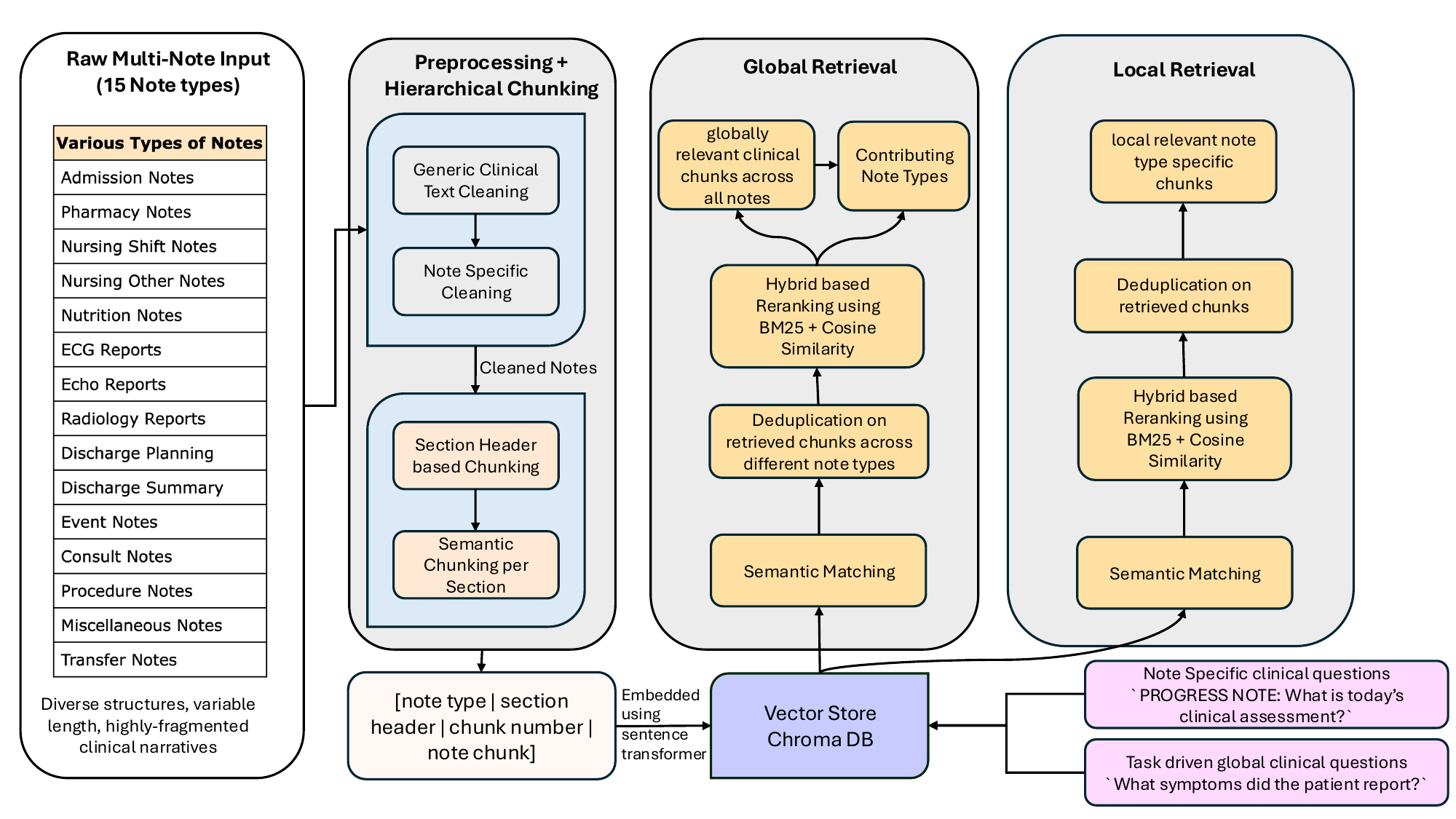}
    \caption{CLI-RAG Architecture: End-to-end flow diagram showing how structured clinical note types are processed including cleaning, chunking, global and local retrieval, and final LLM generation.}
    \label{fig:cli_rag_pipeline}
\end{figure*}
\noindent
\begin{figure*}[t]
    \centering
    \includegraphics[width=1.0\textwidth]{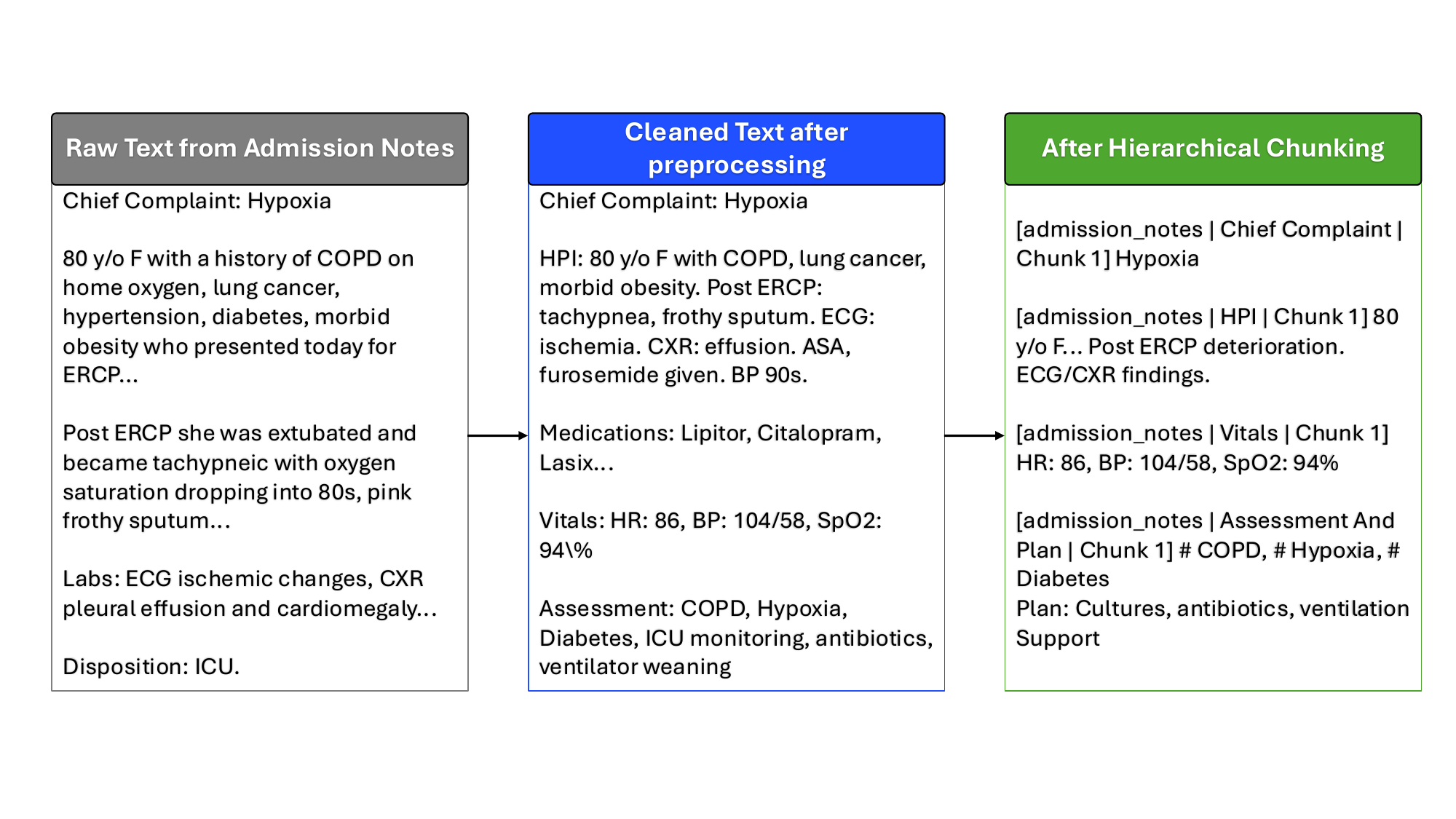}
    \caption{Transformation of a raw free-text note into structured format: Preprocessing eliminates noise, and hierarchical chunking extracts meaningful sections for downstream retrieval.}
    \label{fig:horizontal-cleaning-pipeline}
\end{figure*}
\noindent
\begin{figure*}[t]
    \centering
    \includegraphics[width=1.0\textwidth]{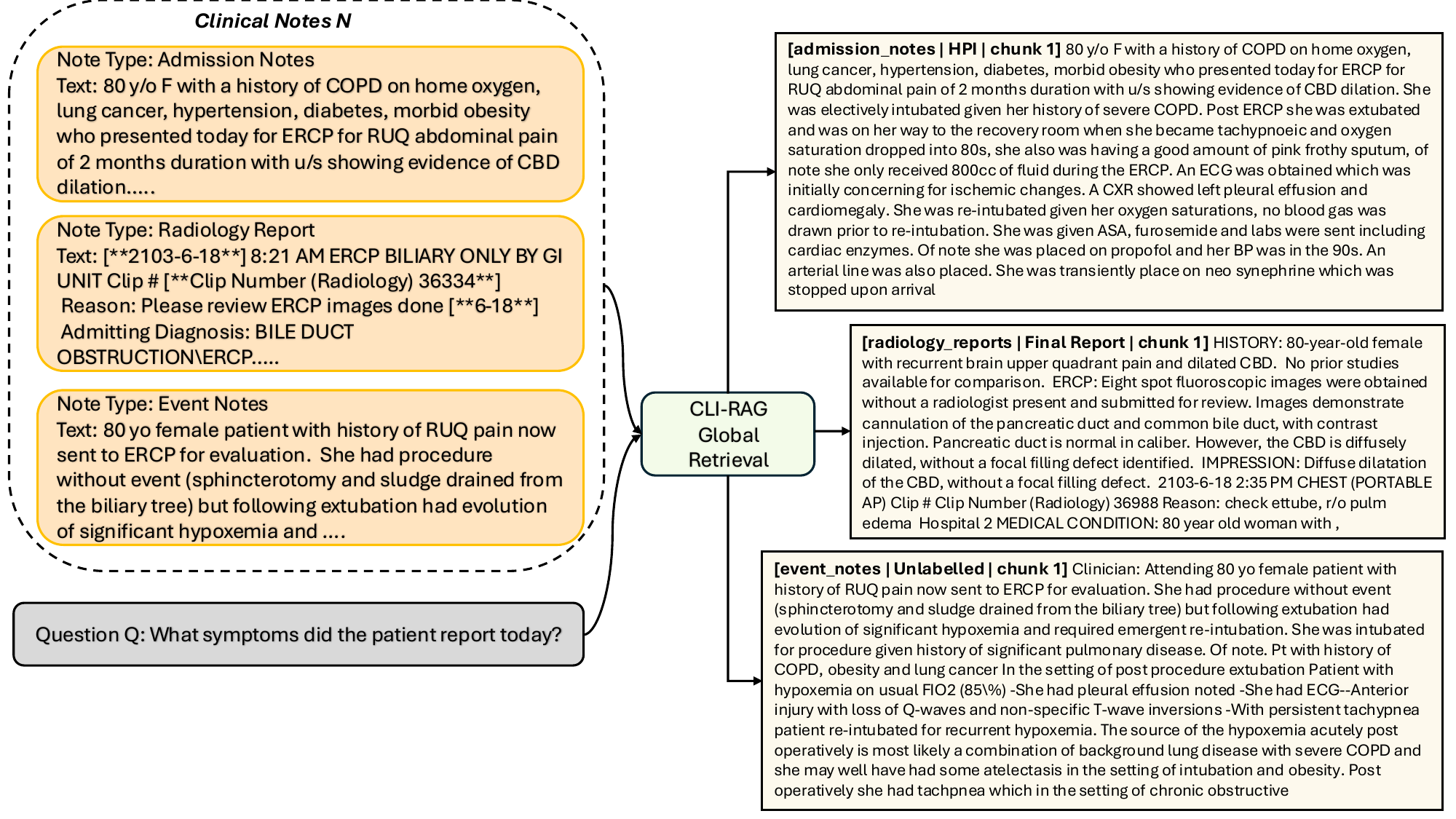}
    \caption{Example of global retrieval in CLI-RAG: Given a clinical question about symptoms, relevant chunks are retrieved from multiple note types (admission, event, radiology) across a patient visit.}
    \label{fig:global-retrieval}
\end{figure*}

\end{document}